# NEW METHOD FOR 3D SHAPE RETRIEVAL

Abdelghni Lakehal and Omar El Beqqali[1]

[1] Sidi Mohamed Ben AbdEllah University GRMS2I FSDM B.P 1796  Fez  Morocco
lakehal_abdelghni@yahoo.fr

## ABSTRACT

*The recent technological progress in acquisition, modeling and processing of 3D data leads to the proliferation of a large number of 3D objects databases. Consequently, the techniques used for content based 3D retrieval has become necessary. In this paper, we introduce a new method for 3D objects recognition and retrieval by using a set of binary images CLI (Characteristic level images). We propose a 3D indexing and search approach based on the similarity between characteristic level images using  Hu moments for it indexing. To measure the similarity between 3D objects we compute the Hausdorff distance between a vectors descriptor. The performance of this new approach is evaluated at set of 3D object of well known database, is NTU (National Taiwan University) database.*

## 1. INTRODUCTION

A number of 3D shape retrieval methods have been proposed. The reader can refer to [1][2]  for a survey of methods. Several methods have been used to characterize the intrinsic attributes, such as the distances to the center [3][4][5], and the curvature [6] of 3D shapes, and to project them onto a sphere to form spherical functions. The spherical harmonics are first introduced in the 3D model retrieval by Vranic et al. in [3] [5].  Tony Tung et al [7] introduce the technique relies upon matching graph representations of 3D-objects, the multiresolutional Reeb graph (MRG) is introduced and used for representing 3D-mesh models and used as a descriptor for 3D object. Vandeborre et al. [8] propose to use full three-dimensional information, the 3D objects are represented as mesh surfaces and 3D shape descriptors are used, the results obtained show the limitation of the approach when the mesh is not regular.

Recent investigations [9] [10][11] illustrate that view-based methods with pose normalization preprocessing get better performance in retrieving rigid models than other approaches and more importantly. *2D view-based* methods [12] [13] consider the 3D shape as a collection of 2D projections taken from canonical viewpoints. Each projection is then described by standard 2D image descriptors like Fourier descriptors [13] or Zernike moments [12]. Chen et al. [14][15] defend the intuitive idea that two 3D models are similar if they also look similar from different angles. Therefore, they use 100 orthogonal projections of an object and encode them by Zernike moments and Fourier descriptors. They also point out that they obtain better results than other well-known descriptors as the MPEG-7 3D Shape Descriptor.  Filali et al.  [16] proposed a framework for 3D models indexing based on 2D views. The goal of their framework is to provide a method for optimal selection of 2D views from a 3D model, and a probabilistic Bayesian method for 3D models indexing from these views.

In the present paper, inspired by the work presented in [9][16][17][18], we introduce a new descriptor for 3D shape. We generate a set of images given by the intersection of the paralleled, specific plans and the 3D shape. We first normalize this shape to guarantee  the invariance of affine transformations, then we extract a number of images called LI (Level Images), after using the K-means method, the number of images is reduced to CLI set (characteristic level images). Each image has been indexed with Hu moments descriptor. The similarity measure between 3D objects returns to measure the similarity between the set of CLI. We will compare the proposed descriptor to two well known descriptors named 3D Zernike moments and 3D surface moments invariants. Then, we will analyze the performance of the proposed descriptor.

This paper is organized as follow: section 2 presents two descriptors with them we will compare our descriptor. The proposed descriptor is presented in the section 3. In section 4, we will see the experimental results. Finally, conclusion summarizes the whole ideas of the present work.

## 2. 3D ZERNIKE MOMENT AND 3D SURFACES MOMENT INVARIANTS

### 2.1 3D Zernike moment descriptor

The 3D Zernike functions $Z_{nl}^m$ are written in Cartesian coordinates [19] using the harmonic polynomials $e_l^m$:

$$Z_{nl}^m(X) = \sum_{\nu=0}^{k} q_{kl}^\nu |X|^{2\nu} e_l^m(X) \qquad (1)$$

While restricting $l$ so that $l \leq n$ and $(n - l)$ be an even number, $2k - n - 1$. And the coefficients $q_{kl}^\nu$ are determined to guarantee the orthonormality. We are now able to define the 3D Zernike moments $\Omega_{nl}^m$ of a 3D object defined by $f$ as

$$\Omega_{nl}^m = \frac{3}{4\pi} \sum_x f(x) \overline{Z_{nl}^m(x)} \qquad (2)$$

Not that the coefficients $Z_{nl}^m$ can be written in a more compact form as a linear combination of monomials of order up to $n$

$$Z_{nl}^m = \sum_{s+r+s \leq n} \chi_{nlm}^{rst} x^r y^s z^t \qquad (3)$$

Finally, the 3D Zernike moments $\Omega_{nl}^m$ of an object can be written as a linear combination of geometrical moments of order up to $n$:

$$\Omega_{nl}^m = \frac{3}{4\pi} \sum_{r+s+t \leq n}' \overline{\chi_{nlm}^{rst}} M_{rst} \qquad (4)$$

Where $M_{rst}$ denotes the geometrical moment of the object scaled to fit in the unit ball:

$$M_{rst} := \sum_{|X| \leq 1} f(X) x^r y^s z^t \qquad (5)$$

Where $X \in \mathbb{R}^3$ denotes the vector $X = (x, y, z)^t$. The collect of the moments into $(2l +1)$-dimensional vectors $\Omega_{nl}^m = (\Omega_{nl}^{-l}, \Omega_{nl}^{-l+1}, \ldots, \Omega_{nl}^{l-1}, \Omega_{nl}^l)$ define the 3D Zernike descriptors $F_{nl}$ as norms of vectors $\Omega_{nl}^m$

$$F_{nl} - \|\Omega_{nl}^m\| \qquad (6)$$

## 2.2 3D Surface moment invariants descriptor

Dong Xu and Hua Li [21] had used a 3-D surface moment invariants as shape descriptors for the representation of free-form surfaces. We consider a 3D surface triangulation $T = \bigcup_{i \in S} T^i$ consisting of triangles $T_i$, $i \in S \subset \mathbb{N}$. The $(k + l + m)^{th}$ order 3D surface moments $M_{klm}$ of $T$ are the accumulated surface moments $m^i_{klm}$, of the associated triangles $T^i$ i.e

$$M_{klm} = \sum_{i \in S} m^i_{klm} \quad (7)$$

For a general triangle $\Delta$ the surface moments are

$$m_{klm} = \iint_{\Delta} x^k y^l z^m \rho(x, y, z) ds \quad (8)$$

with a surface density function $\rho$. Using a surface parameterization $P(u, v) = (x(u, v), y(u, v)z(u, v))$ in $\mathbb{R}^3$, $D$ is definition domain of $(u, v)$ in $\mathbb{R}^2$.The moment (8) can be rewritten as

$$m_{klm} = \iint_D x^k(u, v)y^l(u, v)z^m(u, v)\rho(x(u, v), y(u, v), z(u, v))\sqrt{EG - F^2}du\,dv \quad (9)$$

Where $E = x_u^2 + y_u^2 + z_u^2$ , $G = x_v^2 + y_v^2 + z_v^2$ and $F = x_u x_v + y_u y_v + z_u z_v$ are the coefficients of the first fundamental form.

The centroid of the 3-D surface can be determined from the zero and the first-order moments by $\bar{x} = \frac{M_{100}}{M_{000}}, \bar{y} = \frac{M_{010}}{M_{000}}, \bar{z} = \frac{M_{001}}{M_{000}}$ , then central moment are defined as

$$M_{klm} = \iint_{\Delta} (x - \bar{x})^k (y - \bar{y})^l (z - \bar{z})^m \rho(x, y, z) ds \quad (10)$$

So the central surface moments are invariant under translation. Then, we normalize the surface moments by $M_{000}^{1 + (k + l + m)/2}$, they also became invariant under scaling and can be defined as $\mu_{klm} = \frac{M_{klm}}{M_{000}^{1 + (k + l + m)/2}}$. To construct the surface moments invariant under rotation, D. Xu [21] use four geometric primitives for constructing six invariants consist of 3 fourth order, 2 third order and 1 mixed order surface moment invariants.

# 3. PROPOSED METHOD

We normalize the 3D shape into a canonical coordinate frame, and characterize the shape by a set of characteristic level images noted CLI set. Each one of the CLI set was described by Hu's moments [20]. The final set of moment invariants are used to be the feature vector.

## 3.1 Pose normalization

A 3D object is generally given in arbitrary orientation, scale and position in the 3D space. As most of the 2D/3D shape descriptors do not satisfy the geometrical invariance, pose normalization is then necessary before the 3D object feature extraction. To secure translation, rotation and scale invariance of descriptors, 3D-mesh models are normalized [13][22]. Each triangle mesh model is transformed into a canonical coordinate frame by translating (the center of gravity becomes the origin), rotating (using the Continuous Principal Analysis - CPCA), scaling (the average distance of a point on the surface of the model to the origin becomes 1).

Complete analytical expressions for transforming a mesh model into canonical coordinates are given in [22]. The CPCA is rather efficient and effective for most categories of 3D-objects.

The covariance matrix is calculated, the first eigenvector of this covariance matrix corresponding to the largest eigenvalue points to the direction of the largest variance along which the rotation is applied. When this step has been done, the model is rotated so that the X-axis maps to the eigenvector with the biggest eigenvalue, the Y-axis maps to the eigenvector with the second biggest eigenvalue and the Z-axis maps to the eigenvector with the smallest eigenvalue.

### 3.2 Feature extraction

The method consists on the characterization of 3D objects by a CLI set. Initially, the object 3D has an arbitrary position in the space Figure 1(a), and then has been translated so that its center of mass coincides with the origin, as shown in Figure 1(b), then it is scaled to unit sphere and rotated with the CPCA method Figure 1(c), to alleviate the problem of rotation invariance.

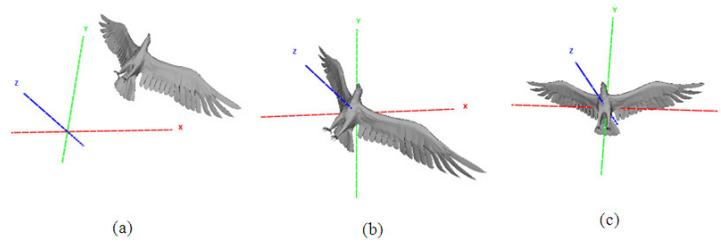

(a)          (b)          (c)

Figure 1. Shape normalization using CCPA

The shape is aligned, Figure 2. (a) (b), so we take 300 perpendicular plan to X-axis, the images generated by the intersection of the shape and this set of plan construct the LI (Level Images) set for a model $M$ of the collection. Denote by $I^M = \{I_1^M, I_2^M, ..., I_N^M\}$ to LI set. Each binary image $I_i^M$ of the model $M$ is generated by the intersection of the shape and the one of perpendicular plane of the X-axes. These plans are equally spaced. For extracting the vector descriptor, among the set of LI, we have to select those that characterize effectively the three-dimensional model to avoid the redundancy problem and to reduce the program complexity. Therefore, we have been applying the K-means technique, so the set $I^M$ had been reduced to the set $I^{CM} = \{I_1^{CM}, I_2^{CM}, ..., I_n^{CM}\}$ that represents the CLI set, Figure 2 (c)(d).

There are two important remarks:

- For each 3D model the number of characteristic level images is fixed in 40 as it shows the Figure 2 (c) (d), the obtained results indicated a promising performance.

- Secondly, the images size varied from on object to the other one, this problem is resolved by using descriptor invariants for 2D binary image.

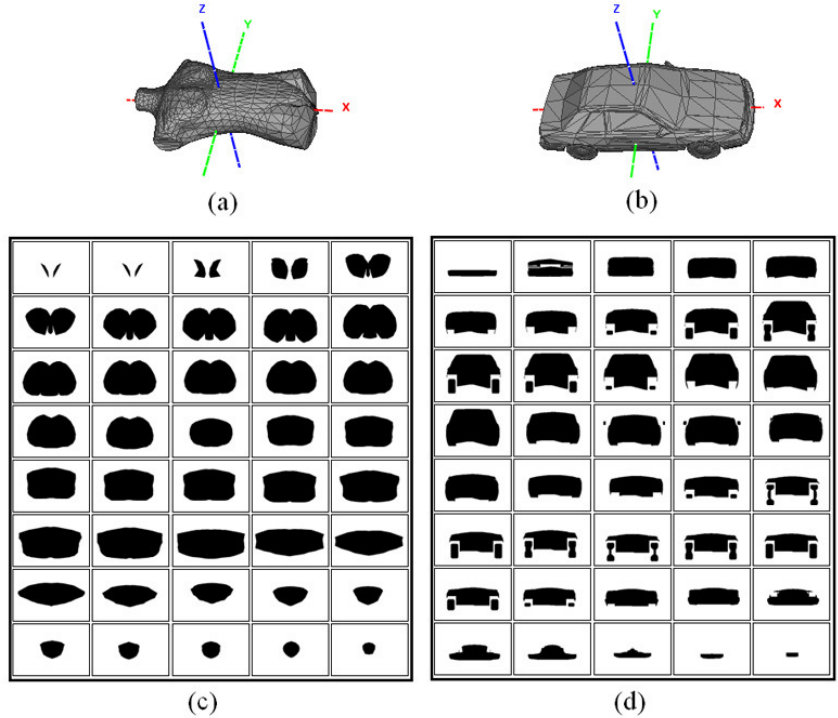

Figure 2. The characteristics level images: (a) and (b) the 3D objects, (c)and (d) their CLI

For each 3D model which are corresponding a number of CLI, the united features of these images also represent the features of the 3D model. The descriptor of the 2D image must be invariant to translation, scaling and rotation. To solve this problem, we use the moments invariant presented by Hu [20]. It is a kind of feature based on region. The 7 moment invariants keep changeless to the translation, rotation and scaling of the object in the images. We can form a feature vector of an image using these 7 moment invariants. The different moments invariants represent different physical meanings.

## 3.2 Similarity measure

In order to measure how similar two objects are, it is necessary to compute distances between pairs of descriptors using a dissimilarity measure. Although the term similarity is often used, dissimilarity corresponds to the notion of distance: small distances means small dissimilarity, and large similarity. A dissimilarity measure can be formalized by a function defined on pairs of descriptors indicating the degree of their resemblance.

Formally speaking, a dissimilarity measure $d$ on a set $S$ is a non-negative valued function:

$$d : S \times S \longrightarrow \mathbb{R}^+$$

$$(x, y) \longrightarrow d(x, y)$$

Given two shapes represented by two sets of points: $X = (X^1, X^2, ..., X^p)$ and $Y = (Y^1, Y^2, ..., Y^q)$, as the number of points of the two set $X$ and $Y$ is not correspondent. The Housdorf distance is commonly used. It is defined as:

$$d_H(X,Y) = \max\left(\max_{x^i \in X} \min_{y^j \in Y} d(X^i, Y^j) \; ; \; \max_{y^j \in Y} \min_{x^i \in X} d(X^i, Y^j)\right)$$

Where $d$ denote the Euclidean distance, is defined as:

$$d(X,Y) = \sqrt{\sum_{i=1}^{n}(X_i - Y_i)^2}$$

Where, $X$ and $Y$ two vectors in $\mathbb{R}^n$.

Each model $M$ of the database is characterized by a vector $X^M = (X_1^M, X_2^M, ..., X_n^M)$, where $X_k^M$ denote to the vector descriptor of the image $I_k^{CM}$ of the model $M$, it is defined as $X_k^M = (X_{k1}^M, X_{k2}^M, ..., X_{k7}^M)$, where $X_{kj}^M$ denote to the j$^{ème}$ Hu moment.

Let $Q$ and $T$ two model of the database, the similarity between them is measured as follow:

$$sim(Q,T) := d_H(X^Q, X^T)$$

Where $d_H(X^Q, X^T) = \max\left\{\max_{i \in S_Q} \min_{j \in S_T} d(X_i^Q, X_j^T), \max_{j \in S_T} \min_{i \in S_Q} d(X_i^Q, X_j^T)\right\}$

$S_M$ denote to the set of indices of the CLI set of the model $M$. And

$$d(X_i^Q, X_j^T) = \sqrt{\sum_{k=1}^{n}(X_{ik}^Q, X_{jk}^T)^2}$$

# 4. EXPERIMENTS AND RESULTS

## 4.1 Test database

Our experiments were performed on a NTU (National Taiwan University) 3D Model Database provides 3D models for research purpose in 3D model retrieval, matching, recognition, classification, clustering and analysis. The database containing 10911 3D models, which are free downloaded from the Internet at [23]. All 3D models are converted into Objet File Format (.off) in the database. In order to evaluate the methods described in this paper, out of these models, 321 were semantically classified into 25 classes, some example of these classes representative are shown in Figure 3.

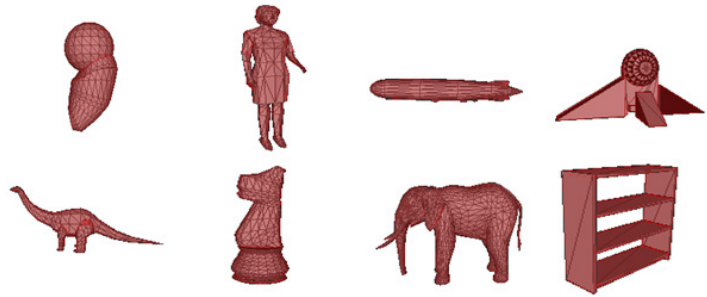

Figure 3. The representatives of some classes of the test database

## 4.2 Results and performance

In order to evaluate the performance of the shape similarity measure, we design experiments performing 3D model retrieval on the test database. Figure 4 shows some of the retrieval examples. We randomly select a model in the database as the query, and then our system returns a list of outputs ranking on the degree of similarity to the input (because the query model in each retrieval test is still from the same test database). We can find that the first several models in the retrieval list are really shape-like to the query model in most cases.

| Query | 1 | 2 | 3 | 4 | 5 |
|-------|---|---|---|---|---|
| 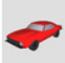 | 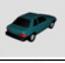 ✓ | 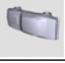 ✓ | 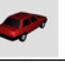 ✓ | 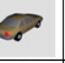 ✓ | 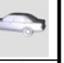 ✓ |
| 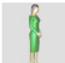 | 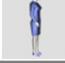 ✓ | 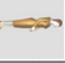 | 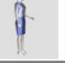 ✓ | 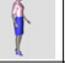 ✓ | 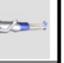 ✓ |
| 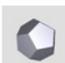 | 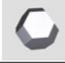 ✓ | 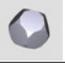 | 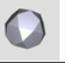 | 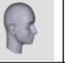 ✗ | 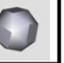 ✓ |
| 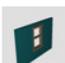 | 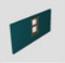 ✓ | 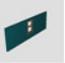 ✓ | 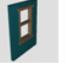 | 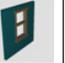 ✓ | 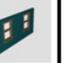 ✓ |
| 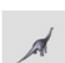 | 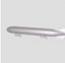 ✗ | 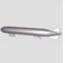 ✗ | 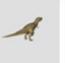 ✓ | 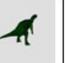 ✓ | 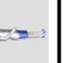 ✗ |

Figure 4.  Some retrieval examples.

In order to evaluate the measurement of retrieval performance we examine Recall-Precision diagram for the shape descriptor of 3D objects. "Precision" measures that the ability of the system to retrieve only models that are relevant while "recall" measures the ability of the system to retrieve all models that are relevant. They are defined as:

$$Recall = \frac{relevant\ correctly\ retrieved}{all\ relevant}$$

$$Precision = \frac{relevant\ correctely\ retrieved}{all\ retrieved}$$

We compared the classification performance of the proposed method with two well known methods, Zernike moments descriptor [19] and surface moments invariants descriptor by using a six invariants [21]. For each one of the chosen shape categories, we have calculated the average Recall-Precision graph by using all shapes of the test database as a query object Figure 5. We can see that the proposed shape descriptors perform better than the 3D Zernike descriptor and surface moments invariants descriptor.

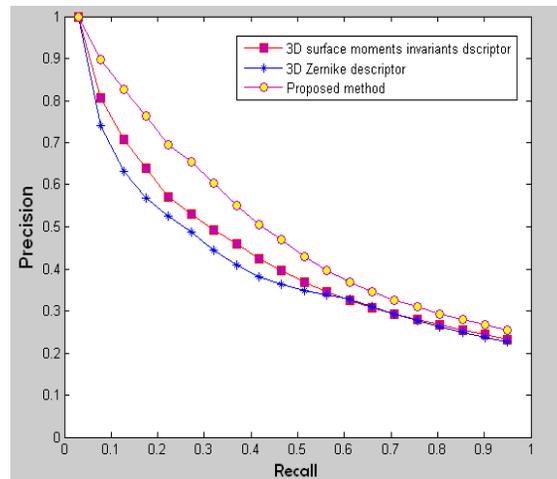

Figure 5. Overall precision-recall graph for all three descriptors

## 5. CONCLUSIONS

In this paper, we have presented a new method for 3D-shape indexing and retrieval, it is based on the 2D *view-based* approach using 2D images for indexing the 3D models. We use a set of binary images called *Characteristic Level images* CLI extracted from the model by the intersection with set of parallel plans. These images are indexed with the 2D images descriptor (we used Hu moments). The similarity between models is calculated by using the Hausdorff. The obtained results show that the proposed descriptor is robust and the comparison with the two will known methods as mentioned above explain good performance.

.

**Authors**

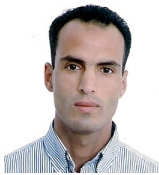

**Lakehal Abdelghni** is a PhD student at the Faculty of Science, Sidi Mohamed Ben Abdellah University, Fez, Morocco. He received his DESA in Scientific Calculate and Optimization and Computer Science from Faculty of Science and Technique in 2006. His current research interests are 2D and 3D shape indexing and retrieval.

**Omar El Beqqali** is currently Professor at Sidi Med Ben AbdEllah University. He is holding a Master in Computer Sciences and a PhD respectively from INSA-Lyon and Claude Bernard University in France. He is leading the 'GRMS2I' research group since 2005 (Information Systems engineering and modeling) of USMBA and the Research-Training PhD Unit 'SM3I'. His main interests nclude Supply Chain field,distributed databases and Pervasive information Systems. He also participated to MED-IST project meetings. O. El Beqqali was visiting professor at UCB-Lyon1 University, INSA-Lyon, Lyon2 University and UIC (University of Illinois of chicago). He is also an editorial board member of the International Journal of Product Lifecycle Management(IJPLM).